\documentclass[letterpaper]{article} 
\usepackage{aaai24}  
\usepackage{times}  
\usepackage{helvet}  
\usepackage{courier}  
\usepackage[hyphens]{url}  
\usepackage{graphicx} 
\usepackage{natbib}  
\usepackage{caption} 
\usepackage{algorithm}
\usepackage{algorithmic}
\usepackage{amsmath}
\usepackage{amssymb}
\usepackage{paralist}
\usepackage{array}
\usepackage{booktabs}
\usepackage{multirow}
\usepackage{bm}
\usepackage{diagbox}
\usepackage{makecell}
\usepackage{color, colortbl}
\definecolor{Gray}{gray}{0.9}
\newcommand{\figref}[1]{Fig.~\ref{#1}}%
\newcommand{\tabref}[1]{Table~\ref{#1}}%

\renewcommand{\eqref}[1]{Eq.~(\ref{#1})}
\renewcommand\arraystretch{1.2}

\usepackage{newfloat}
\usepackage{listings}
\DeclareCaptionStyle{ruled}{labelfont=normalfont,labelsep=colon,strut=off} 
\floatstyle{ruled}
\newfloat{listing}{tb}{lst}{}
\floatname{listing}{Listing}
\title{BEV-MAE: Bird's Eye View Masked Autoencoders for Point Cloud \\ Pre-training in Autonomous Driving Scenarios}
\author{
    Zhiwei Lin\textsuperscript{\rm 1},
    Yongtao Wang\textsuperscript{\rm 1}\thanks{Corresponding author.},
    Shengxiang Qi\textsuperscript{\rm 2},
    Nan Dong\textsuperscript{\rm 2},
    Ming-Hsuan Yang\textsuperscript{\rm 3} \\
}
\affiliations{
    \textsuperscript{\rm 1}Wangxuan Institute of Computer Technology, Peking University\\
    \textsuperscript{\rm 2}Chongqing Changan Automobile Co., Ltd\\
    \textsuperscript{\rm 3}University of California, Merced\\

    \{zwlin, wyt\}@pku.edu.cn, 
    shengxiang.qi@gmail.com,
    dongnan1@changan.com.cn,
    mhyang@ucmerced.edu
}

\begin{document}

\maketitle

\begin{abstract}
Existing LiDAR-based 3D object detection methods for autonomous driving scenarios mainly adopt the training-from-scratch paradigm. 
Unfortunately, this paradigm heavily relies on large-scale labeled data, whose collection can be expensive and time-consuming.
Self-supervised pre-training is an effective and desirable way to alleviate this dependence on extensive annotated data.
In this work, we present BEV-MAE, an efficient masked autoencoder pre-training framework for LiDAR-based 3D object detection in autonomous driving.
Specifically, we propose a bird's eye view (BEV) guided masking strategy to guide the 3D encoder learning feature representation in a BEV perspective and avoid complex decoder design during pre-training.
Furthermore, we introduce a learnable point token to maintain a consistent receptive field size of the 3D encoder with fine-tuning for masked point cloud inputs.
Based on the property of outdoor point clouds in autonomous driving scenarios, \textit{i.e.}, the point clouds of distant objects are more sparse, we propose point density prediction to enable the 3D encoder to learn location information, which is essential for object detection.
Experimental results show that BEV-MAE surpasses prior state-of-the-art self-supervised methods and achieves a favorably pre-training efficiency.
Furthermore, based on TransFusion-L, BEV-MAE achieves new state-of-the-art LiDAR-based 3D object detection results, with 73.6 NDS and 69.6 mAP on the nuScenes benchmark.
%
The source code will be released at \url{https://github.com/VDIGPKU/BEV-MAE}.
\end{abstract}

\section{Introduction}
\label{sec:intro}
\begin{figure}[!t]
  \centering
	\includegraphics[width=0.95\linewidth]{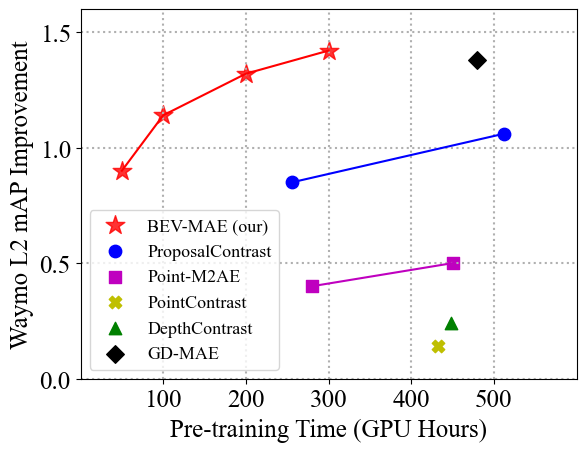}
  \caption{\textbf{Performance improvement \textit{vs.} Pre-training time trade-off.} All entries are benchmarked by a P40 GPU. The 3D object detector is CenterPoint~\cite{yin2020center}. All models are pre-trained on full Waymo and then fine-tuned with 20\% training samples on Waymo. 
  }
    \label{fig:gpu}
\end{figure}

3D object detection is one of the most basic tasks in autonomous driving.
It aims to localize objects in 3D space and classify them simultaneously. 
In recent years, LiDAR-based 3D object detection methods have achieved significant success due to the increasing amount of labeled training data~\cite{nuscenes}. 
However, existing LiDAR-based 3D object detection methods for autonomous driving scenarios often adopt the paradigm of training from scratch, which brings two defects.
First, the training-from-scratch paradigm largely relies on extensive labeled data. 
For 3D object detection, annotating precise bounding boxes and classification labels is costly and time-consuming, \textit{e.g.}, it takes around 114$s$ to annotate one object~\cite{meng2020weakly} on KITTI.
Second, self-driving vehicles can generate massive unlabeled point cloud data daily in many practical scenarios, which cannot be used in the training-from-scratch paradigm.

\begin{figure}[!t]
  \centering
	\includegraphics[width=0.95\linewidth]{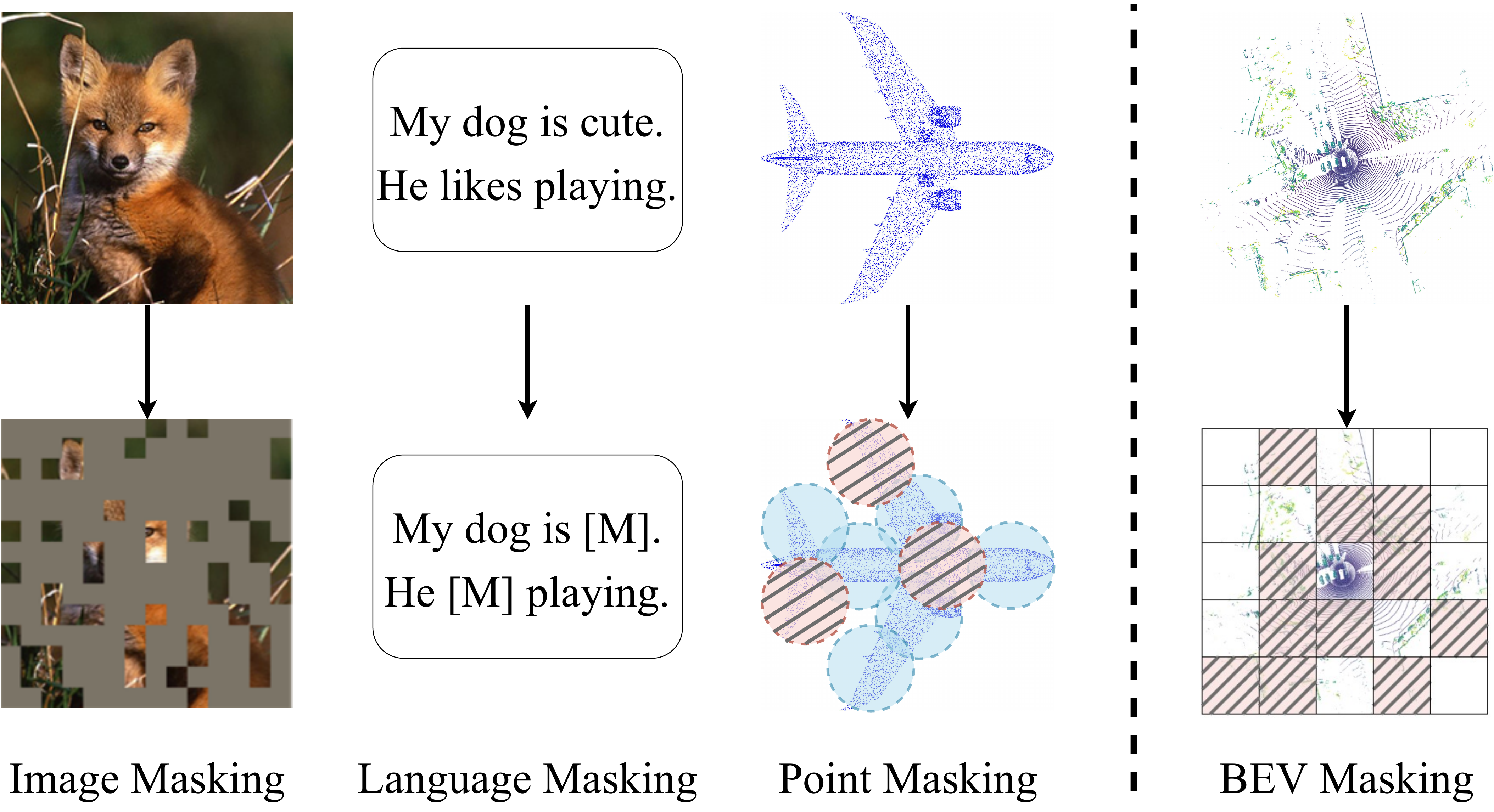}
  \caption{\textbf{Illustration of several masking strategies in the masked modeling.}
  MAE~\cite{DBLP:conf/cvpr/mae} masks non-overlapping image patches. BERT~\cite{devlin2018bert} masks words or sentences. Point-MAE~\cite{DBLP:journals/corr/pointmae} uses furthest point sampling to create overlapping point patches.
  Our method (right) projects point clouds into a BEV plane, and masks points in non-overlapping BEV grids. 
  }
    \label{fig:illustration}
\end{figure}

A simple and desirable solution to the above two problems is self-supervised pre-training, widely used in fields such as computer vision~\cite{DBLP:conf/cvpr/mae} and natural language processing~\cite{devlin2018bert}. 
By solving pre-designed pretext tasks, self-supervised pre-training can learn a general and transferable feature representation on large-scale unlabeled data.
One of the mainstream approaches in self-supervised learning is masked modeling, as illustrated in \figref{fig:illustration}.
Specifically, numerous approaches adopt masked image modeling~\cite{DBLP:conf/cvpr/mae,DBLP:conf/cvpr/simmim} and masked language modeling~\cite{devlin2018bert} to pre-train networks by reconstructing images, words, and sentences from masked inputs. 
Recently, several methods~\cite{DBLP:conf/cvpr/pointbert, DBLP:journals/corr/pointmae} use masked point modeling for dense point clouds, achieving promising performance on shape classification, shape segmentation, and indoor 3D object detection.
However, these schemes mainly focus on synthetic or indoor datasets, such as ShapeNet~\cite{DBLP:journals/corr/shapenet}, ModelNet40~\cite{DBLP:conf/cvpr/modelnet}, and ScanNet~\cite{dai2017scannet}.
When applied to autonomous driving scenarios, where the range of scenes is more extensive and point cloud density is more sparse, their results are unsatisfactory~\cite{liang2021exploring}.
To address this issue, a few works~\cite{voxelmae, geomae} explore masked modeling for autonomous driving scenarios.
However, these methods mainly adopt voxel-based masking strategies and recover masked points in voxels.
There exist feature representation gaps between these \textit{voxel-based} pre-training pipelines and the prevalent \textit{BEV-based} 3D object detection methods.

In this work, we present bird's eye view masked autoencoders, dubbed BEV-MAE, specially for pre-training 3D object detectors on autonomous driving point clouds. 
Instead of randomly masking point clouds or voxels, we propose a BEV-guided masking strategy (right part of \figref{fig:illustration}) for two benefits.
First, we enforce the 3D encoder to learn feature representation in a BEV perspective by reconstructing masked information on the BEV plane.
Therefore, during fine-tuning, the pre-trained 3D encoder can facilitate the training process of 3D detectors in the BEV perspective.
Second, current 3D encoders of LiDAR-based 3D object detectors often downsample the resolution of points or voxels to save the computational overhead, \textit{e.g.}, GPU memory, and training time. 
With the BEV-guided masking strategy, we do not need to design complicated decoders with upsampling operations~\cite{shi2020unet} since the size of masked girds matches the resolution of BEV features.
BEV-MAE can achieve promising results with a simple one-layer 3$\times$3 convolution as the decoder.
By designing the BEV-guided masking strategy, we obtain considerable pre-training efficiency improvement, as shown in the \figref{fig:gpu}. 
%

Since the commonly used sparse 3D convolutions only perform computation near the occupied areas, the receptive field size of the 3D encoder may become smaller with masked point inputs, resulting in low learning efficiency and poor transferability.
To alleviate this issue, we replace the masked points with a shared learnable point token during pre-training to maintain a consistent receptive field size of the 3D encoder with fine-tuning.
The shared learnable point token can help communication between unmasked areas without providing additional information to reduce the difficulty of the pre-training task.
Moreover, we introduce point density prediction for masked areas in addition to using coordinates of masked point clouds as the reconstruction target~\cite{voxelmae, gdmae}.
Since the point clouds become sparse when they are far from the LiDAR sensor in outdoor scenes, the density of point clouds can reflect the distance between points and the central LiDAR sensor.
Naturally, density prediction can guide models to learn location information, which is critical for object detection.

The main contributions of this work are:
    \begin{compactitem}
        \item We present a simple and efficient self-supervised pre-training method, BEV-MAE, tailored for LiDAR-based 3D object detectors in autonomous driving. 
        With the proposed BEV-guided masking strategy, the 3D encoder of 3D object detectors can directly learn feature representation in a BEV perspective with a simple and lightweight decoder.
        \item We introduce a shared learnable point token to alleviate the inconsistency of the receptive field size for the 3D encoder during pre-training and fine-tuning, and propose density prediction to learn location information for the 3D encoder.
        \item BEV-MAE outperforms existing self-supervised methods in performance and pre-training efficiency. 
        Moreover, BEV-MAE can further improve the performance of state-of-the-art 3D object detectors. Combined with TransFusion-L, BEV-MAE achieves new state-of-the-art results on nuScenes.
    \end{compactitem}

\begin{figure*}[!t]
  \centering
	\includegraphics[width=0.97\linewidth]{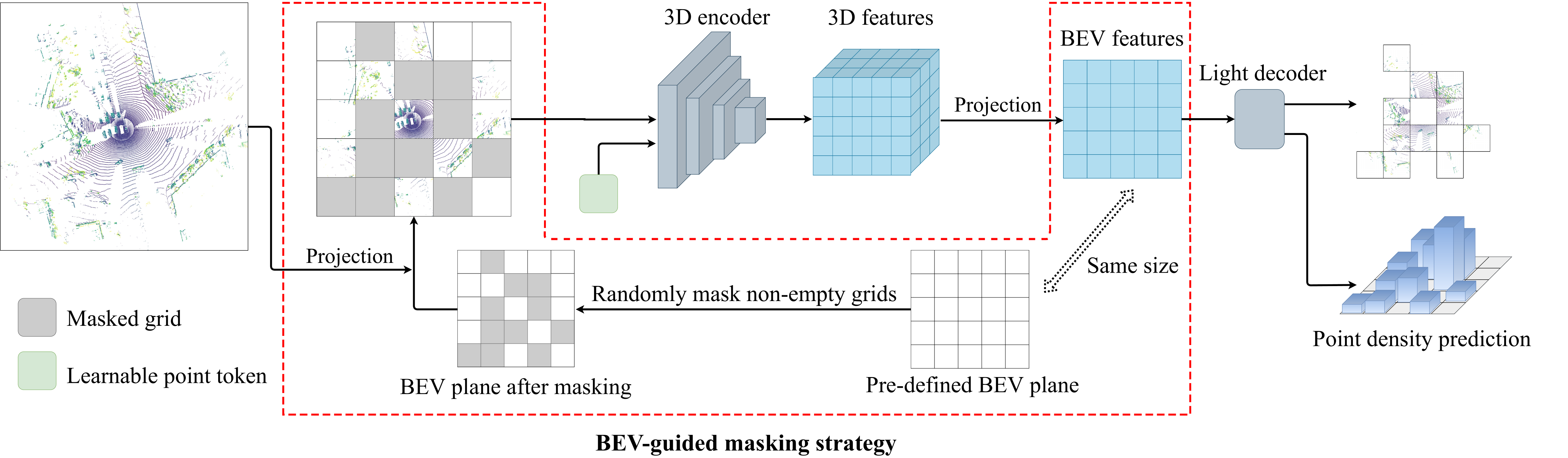}
  \caption{\textbf{Overall pipeline of BEV-MAE.} We first mask point clouds with the BEV-guided masking strategy. Then, the masked points are replaced with a shared learnable point token. 
After extracting BEV features by a 3D encoder from visible points, we send the features to a light decoder to reconstruct masked point clouds and predict the point density of masked grids.
  }
    \label{fig:pipeline}
\end{figure*}

\section{Related Work}
\label{sec:related work}
\subsection{3D Object Detection}
The objective of 3D object detection is to localize objects of interest with 3D bounding boxes and classify the detected objects.
Due to the large domain gap between indoor and outdoor datasets, the two cases' corresponding 3D object detection methods have been developing almost independently.
Here we focus on the works about outdoor 3D object detection for autonomous driving.
Recent outdoor 3D object detection approaches~\cite{yan2018second, yin2020center, lang2019pointpillars} based on BEV representation attract much attention for their convenience in fusing different information, including multi-view~\cite{DBLP:journals/corr/bevdet}, multi-modality~\cite{DBLP:journals/corr/bevfusion}, and temporal inputs~\cite{DBLP:journals/corr/bevdet4d}.
To obtain BEV features, LiDAR-based methods first extract point features by a 3D encoder, such as PointPillar~\cite{lang2019pointpillars} and VoxelNet~\cite{DBLP:conf/cvpr/voxelnet}, and then project the features onto a BEV plane according to the point coordinates. 
For camera-based approaches, they extract the 2D features from multi-view images by a prevalent 2D backbone, including ResNet~\cite{he2016deep} and SwinTransformer~\cite{liu2021swin}, and utilize geometry-based view transformation~\cite{DBLP:conf/eccv/lift} to construct BEV features from multi-view image features.
Next, a 2D encoder and a detection head~\cite{yan2018second, yin2020center} are applied for these methods to process the BEV features and predict the final detection results.
In essence, this pipeline reduces the task of 3D object detection to 2D object detection.
Hence, these methods can fully utilize the highly developed 2D object detection algorithms.
Similarly, our work pre-trains the 3D encoder of 3D object detectors in the BEV space and can directly facilitate the 3D detection task in the BEV perspective. 

\subsection{Self-supervised Learning for Outdoor Point Clouds}
The main idea of self-supervised learning is to train networks on unlabeled data with pretext tasks. 
Since no prior human knowledge is introduced, the pre-trained networks can learn a more general feature representation and have an excellent transfer ability.
Recently, 3D representation learning of outdoor point clouds has attempted to adopt the core idea of self-supervised learning. 
GCC-3D~\cite{liang2021exploring} proposes geometry-aware contrast and harmonized cluster to learn geometry and semantic information from sparse point clouds.
ProposalConstrast~\cite{DBLP:conf/eccv/proposal} samples region proposals for each point cloud with farthest point sampling (FPS) and jointly optimizes inter-proposal discrimination and inter-cluster separation.
In addition to contrastive learning methods, masked modeling for outdoor point clouds has been studied recently for its simplicity and efficiency.
Voxel-MAE~\cite{voxelmae} proposes a voxel-wised masking strategy for the transformer-based encoder. It reconstructs masked voxels and predicts the number of points and occupancy.
GD-MAE~\cite{gdmae} presents a multi-level transformer architecture and adopts a multi-scale masking strategy. It uses a generative decoder to recover masked patches with multi-scale features.
GeoMAE~\cite{geomae} introduces pyramid centroid, occupancy, surface normal, and surface curvature of point clouds as prediction targets.
MV-JAR~\cite{mvjar} proposes a masked voxel Jigsaw and reconstruction method.
MSP~\cite{msp} builds a masked shape prediction pipeline for 3D scene understanding.
ALSO~\cite{also} presents a query-based occupancy estimation to capture 3D semantic information.
However, these works mainly adopt voxel-based masking strategies.

This paper proposes an efficient masked modeling framework, BEV-MAE, for pre-training point clouds in autonomous driving scenarios. 
Instead of masking voxels, BEV-MAE adopts a BEV-guided masking strategy to learn BEV feature representation and achieves new state-of-the-art self-supervised learning performance.

\section{Method}
The overall pipeline of BEV-MAE is shown in \figref{fig:pipeline}. 
BEV-MAE first uses a BEV-guided masking strategy to mask point clouds. 
These masked points are then replaced with a shared learnable point token. 
We send the processed points into a 3D encoder and a light decoder sequentially.
Finally, the light decoder will reconstruct the masked points and predict the point density of the masked area.

\subsection{BEV-guided Masking Strategy}
In LiDAR-based 3D object detection, the point clouds are often divided into regular voxels. 
A straightforward masking strategy is to mask the voxels like masking patches in vision~\cite{DBLP:conf/cvpr/mae, DBLP:conf/cvpr/simmim}. 
However, this simple voxel masking strategy makes subsequent decoder design difficult. 
In addition, it takes little consideration of the type of feature representation in the mainstream 3D object detection methods for autonomous driving, \textit{i.e.}, BEV feature representation.
To this end, we propose a BEV-guided masking strategy to mask points in the BEV plane. 

Specifically, assuming the resolution of the features in the BEV perspective after encoding and transformation is $X\times Y\times C$, we first pre-define a grid-shaped BEV plane with the size of $X\times Y$. 
We then project each LiDAR point $p_k$ into a corresponding BEV grid $g_{i,j}$ of the pre-defined BEV plane according to its point coordinates $(x_{p_k},y_{p_k})$.
Each BEV grid will contain a various number of points:
\begin{equation}
    g_{i,j} = \{p_k~|~ \lfloor x_{p_k}/d \rfloor = i,~\lfloor y_{p_k}/d \rfloor = j\},
\end{equation}
where $d$ is the downsample ratio of the 3D encoder and $\lfloor x \rfloor$ denotes rounding down of $x$.
We randomly select a significant fraction of non-empty BEV grids, \textit{i.e.}, $g_{i,j}\neq \varnothing$, as the masked grids $\{g^m_i\}$ and denote the remaining BEV grids as visible grids $\{g^v_i\}$.
Finally, we obtain the visible and masked point clouds by merging the points in $\{g^m_i\}$ and $\{g^v_i\}$, formulated as $\{p_k^v\} = \mathop{\cup}\limits_{\tiny{i}} g^v_i$ and $\{p_k^m\} = \mathop{\cup}\limits_{\tiny{i}} g^m_i$ respectively.

\subsection{Learnable Point Token}
The 3D encoder of recent voxel-based 3D object detectors typically consists of several sparse convolution operations, which only process the features near the non-empty voxels. 
When only taking visible point clouds $\{p_k^v\}$ as input, the size of the receptive field of the 3D encoder becomes smaller. 
To address this issue, we replace the masked point clouds $\{p_k^m\}$ with a shared learnable point token. 
Specifically, we use coordinates of full point clouds as the input indexes~\cite{yan2018second} of sparse convolution and replace the feature of masked point clouds with the shared learnable point token in the first sparse convolution layer.
We keep the other sparse convolutional layers unchanged.
The goal of the proposed shared learnable point token is to pass the information from one point or voxel to another to maintain the size of the receptive field. 
It does not introduce any additional information, including the coordinates of masked points.

\subsection{Decoder Design}
In the BEV-guided masking strategy, the size of masked areas is aligned with the resolution of the BEV features.
Thus, we can directly predict the reconstruction results of one masked grid from the corresponding BEV features without upsampling operations.

The decoder of BEV-MAE is only used during pre-training to solve the masking task. 
Naturally, the design of the decoder is flexible and independent of the 3D encoder architecture.
We evaluate three types of decoders, \textit{i.e.}, one-layer 3$\times$3 convolution, transformer block~\cite{DBLP:conf/cvpr/mae}, and residual convolution block~\cite{he2016deep}, and find the highly lightweight decoder, a one-layer 3$\times$3 convolution, can achieve impressive performance while reducing pre-training time and GPU memory cost.

\subsection{Reconstruction Target}
The proposed BEV-MAE is supervised by two tasks, \textit{i.e.}, point cloud reconstruction, and density prediction. 
A separate linear layer is applied as the prediction head for each task to predict results. 
We describe each task below.

\subsubsection{Point Cloud Reconstruction.}
BEV-MAE reconstructs the masked input by predicting the coordinates of masked points. 
However, each masked grid in the pre-defined BEV plane contains a different number of points, leading to the challenges of designing the prediction head.

To address this issue, we propose to reconstruct the local structure of the masked point clouds with a set-to-set prediction.
Specifically, we apply a linear layer to predict a set of 3D points with a fixed number, denoted as $P_i=\{p_l~|~l=1,2,..., L\}$, for each masked grid $g_i^m$.
Given the original points $\hat{P}_i=\{\hat{p_k}~|~k=1,2,..., N\}$ in $g_i^m$, where $N$ varies with the grid, we utilize Chamfer Distance~\cite{DBLP:conf/cvpr/chamfer} between predictions $P_i$ and the ground-truth $\hat{P}_i$ as the reconstruction loss:
\begin{equation}
    \mathcal{L}_c^i = \frac{1}{L}\sum_{p_l\in P_i}{\min_{\hat{p_k}\in \hat{P_i}} \lvert\lvert p_l-\hat{p_k} \rvert\rvert_2^2} + \frac{1}{N}\sum_{\hat{p_k}\in \hat{P_i}}{\min_{p_l\in P_i} \lvert\lvert \hat{p_k}-p_l \rvert\rvert_2^2}.
\end{equation}
We then average the loss over all the masked grids as the final reconstruction loss:
\begin{equation}
    \mathcal{L}_c = \frac{1}{n_m} \sum_{i=1}^{n_m}\mathcal{L}_c^i,
\end{equation}
where $n_m$ is the number of masked grids.
The Chamfer distance measures the distance between two sets with different cardinalities.
Thus, it enforces the shape of predicted point clouds to mimic the local structure of masked inputs~\cite{DBLP:conf/cvpr/chamfer}.

Since the coordinate value of each ground-truth point varies largely in the different grids, directly predicting the absolute coordinates of each point may cause instability during pre-training.
To alleviate this issue, we predict the normalized coordinate in point cloud reconstruction.
Specifically, we first calculate the coordinate offset of each ground-truth point to the center of its corresponding BEV grid and then normalize the offset value by the size of the BEV grid.

\setlength\tabcolsep{.2em}
\begin{table*}[!t]
\begin{center}
\small
\begin{tabular}{l|c|c|c|c|ccc}
\Xhline{0.8px}
\multicolumn{1}{c|}{\multirow{2}{*}[-0.5ex]{{Pre-traing Method}}} & 
\multirow{2}{*}[-0.5ex]{\shortstack{Epochs}} 
& 
\multirow{2}{*}[-0.5ex]{\shortstack{Time}} 
&
\multirow{2}{*}[-0.5ex]{\shortstack{Dataset \\ fraction}} 
& \multicolumn{4}{c}{L2 (mAP/APH)} \\
\cline{5-8}
& & & & Overall & Vehicle & Pedestrian & Cyclist \\
\Xhline{0.6px}
From-scratch & -&- & -& 65.60 / 63.21 & 64.18 / 63.69 & 65.22 / 59.68 & 67.41 / 66.25  \\
GCC-3D~\cite{liang2021exploring}$^*$ & 40 & - & 100\% & 65.29 / 62.79 & 63.97 / 63.47 & 64.23 / 58.47 & 67.68 / 66.44 \\
PointContrast~\cite{xie2020pointcontrast} & 50 & 54h & 100\% & 65.88${^{+0.28}}$ / 63.28${^{+0.07}}$ & 63.81 / 63.33 & 66.67 / 60.51 & 67.17 / 66.00 \\
DepthContrast~\cite{DBLP:conf/iccv/depthcontrast} & 50 & 56h & 100\% & 65.84${^{+0.24}}$ / 63.33${^{+0.12}}$ & 64.45 / 63.95 & 65.61 / 59.86 & 67.43 / 66.22 \\
Point-M2AE~\cite{DBLP:journals/corr/m2ae} & 30 & 56h & 100\% & 66.10${^{+0.50}}$ / 63.59${^{+0.38}}$ & 64.26 / 63.77 & 65.64 / 60.00 & 68.20 / 67.01 \\
ProposalContrast~\cite{DBLP:conf/eccv/proposal} & 50 & 64h & 100\% & 66.42${^{+0.82}}$ / 63.85${^{+0.64}}$ & 65.03 / 64.53 & 65.93 / 59.95 & 68.26 / 67.04 \\
MSP~\cite{msp} & 30 & - & 100\% &~~~~~~~~~~~~~~~ - / 64.26${^{+1.05}}$ & - / - & - / - & - / - \\
GD-MAE\textsuperscript{$\dagger$}~\cite{gdmae} & 30 & 60h & 100\% & 66.98${^{+1.38}}$ / 64.53${^{+1.32}}$ & 65.64 / 64.95 & 66.39 / 61.12 & 68.92 / 67.52 \\
\Xhline{0.6px}
BEV-MAE (Ours) & 20 & 5h & 20\% & 66.70${^{+1.10}}$ / 64.25${^{+1.04}}$ & 64.71 / 64.22 & 66.21 / 60.59 & 69.11 / 67.93\\
BEV-MAE (Ours) & 30 & 38h & 100\% & \textbf{67.02}${\bm{^{+1.42}}}$ \textbf{/} \textbf{64.55}${\bm{^{+1.34}}}$ & {65.01 / 64.53} & {66.58 / 60.87} & {69.46 / 68.25} \\
\Xhline{0.8px}
\end{tabular}
\caption{{Comparisons between BEV-MAE and state-of-the-art self-supervised learning methods on Waymo validation set.}
All detectors are fine-tuning with 20\% training samples on Waymo following the OpenPCDet configuration. 
Here, the entry with $^{*}$ denotes the results are from the paper~\cite{liang2021exploring};
the entry with $\dagger$ indicates the results are implemented by the released official code$^{1}$.
`Epochs' indicates the pre-training epochs;
`Dataset fraction' means the data fraction of the Waymo training set used for pre-training;
and `Time' refers to the pre-training time estimated by 8 P40 GPU.
}
\label{tab:main result waymo}
\end{center}
\end{table*}

\begin{table}[!t]
\begin{center}
\small
\begin{tabular}{l|cc}
\Xhline{0.8px}
Method & NDS & mAP \\
\Xhline{0.6px}
CenterPoint~\cite{yin2020center} & 65.5 &58.0  \\
VISTA~\cite{deng2022vista}  &69.8 &63.0 \\
FocalsConv~\cite{focalsconv-chen}  &70.0 &63.8 \\
VoxelNeXt~\cite{voxelnext}  & 70.0& 64.5 \\
TransFusion-L~\cite{bai2022transfusion} &70.2&65.5\\
LargeKernel3D~\cite{largekernel3d} & 70.6 & 65.4\\
Link~\cite{link} & 71.0  & 66.9\\
GeoMAE$^{\dagger}$\cite{geomae}  & 72.5 & 67.8 \\
\Xhline{0.6px}
BEV-MAE (Ours)  & 71.7 & 67.0 \\
BEV-MAE$^{\dagger}$ (Ours)  & \textbf{73.6} & \textbf{69.6} \\

\Xhline{0.8px}
\end{tabular}
\caption{{Performances of 3D object detection on the nuScenes \textit{test}
split.} ${\dagger}$ results are obtained using a modified model structure.
}
\label{tab:main result nus}
\end{center}
\end{table}

\subsubsection{Point Density Prediction.}
Unlike image, language, and indoor point clouds, outdoor point clouds in autonomous driving scenarios have the property that the density of the point clouds becomes small when they are far from the LiDAR sensor. 
Consequently, the density can reflect the location of each point or object.
Furthermore, for object detection, the localization ability of detectors is essential.
Based on the above analysis, we propose another task for BEV-MAE, \textit{i.e.}, point density prediction, to guide the 3D encoder to achieve a better localization ability.
Compared with predicting the number of points in Voxel-MAE~\cite{voxelmae}, our proposed density prediction is more stable and effective during training.

For each masked grid $g_i^m$, we count the number of points in this grid and calculate the density $\hat{\rho_i}$ by dividing the number of points with the occupied volume in 3D space as the ground truth for density prediction. 
Then, we use a linear layer as the prediction head to obtain the density prediction $\rho_i$.
We supervise this task with the \textit{Smooth}-$\ell_1$ loss:
\begin{equation}
    \mathcal{L}_d^i = Smooth\text{-}{\ell_1}(\rho_i-\hat{\rho_i}).
\end{equation}
Similarly, we average the loss over all the masked grids as the final density prediction loss:
\begin{equation}
    \mathcal{L}_d = \frac{1}{n_m} \sum_{i=1}^{n_m}\mathcal{L}_d^i.
\end{equation}

\newcolumntype{g}{>{\columncolor{Gray}}c}
\setlength\tabcolsep{.2em}
\begin{table*}[!t]
\centering
\small
\begin{tabular}{c|c|cc|cccccc}
\Xhline{0.8px}
\multirow{2}{*}{Data amount} & \multirow{2}{*}{Initialization} & \multicolumn{2}{c|}{Overall} & \multicolumn{2}{c}{Car} & \multicolumn{2}{c}{Pedestrian} & \multicolumn{2}{c}{Cyclist} \\ 
\cline{3-10}
 &  & L2 mAP & L2 mAPH & L2 mAP & L2 mAPH & L2 mAP & L2 mAPH & L2 mAP & L2 mAPH 
 \\ \Xhline{0.6px}
\multirow{5}{*}{5\%} & Random & 44.41 & 40.34 & 51.01 & 50.49 & 52.74 & 42.26 & 29.49 & 28.27 \\
 & PointContrast\cite{xie2020pointcontrast} & 45.32 & 41.30 & 52.12 & 51.61 & 53.68 & 43.22 & 30.16 & 29.09 \\
 & ProposalContrast\cite{DBLP:conf/eccv/proposal} & 46.62 & 42.58 & 52.67 & 52.19 & 54.31 & 43.82 & 32.87 & 31.72 \\
 & MV-JAR~\cite{mvjar} & {50.52} & {46.68} & 56.47 & 56.01 & 57.65 & 47.69 & 37.44 & 36.33 \\
 &\cellcolor{Gray} BEV-MAE (Ours) &\cellcolor{Gray} \textbf{51.63}&\cellcolor{Gray}\textbf{47.77} &\cellcolor{Gray} {56.35}&\cellcolor{Gray}{55.81} &\cellcolor{Gray} {58.11}&\cellcolor{Gray}{48.37} &\cellcolor{Gray} {40.44}&\cellcolor{Gray}{39.13} \\
 \Xhline{0.6px}
\multirow{5}{*}{10\%} & Random & 54.31 & 50.46 & 54.84 & 54.37 & 60.55 & 50.71 & 47.55 & 46.29 \\
 & PointContrast\cite{xie2020pointcontrast} & 53.69 & 49.94 & 54.76 & 54.30 & 59.75 & 50.12 & 46.57 & 45.39 \\
 & ProposalContrast\cite{DBLP:conf/eccv/proposal} & 53.89 & 50.13 & 55.18 & 54.71 & 60.01 & 50.39 & 46.48 & 45.28 \\
 & MV-JAR~\cite{mvjar} & {57.44} & {54.06} & 58.43 & 58.00 & 63.28 & 54.66 & 50.63 & 49.52 \\
 & \cellcolor{Gray}BEV-MAE (Ours) &\cellcolor{Gray} \textbf{58.16}&\cellcolor{Gray}\textbf{54.75} &\cellcolor{Gray} 58.51&\cellcolor{Gray}57.94 &\cellcolor{Gray} 63.83&\cellcolor{Gray}55.23 &\cellcolor{Gray}52.13&\cellcolor{Gray}51.07 \\
 \Xhline{0.6px}
\multirow{5}{*}{20\%} & Random & 60.16 & 56.78 & 58.79 & 58.35 & 65.63 & 57.04 & 56.07 & 54.94 \\
 & PointContrast\cite{xie2020pointcontrast} & 59.35 & 55.78 & 58.64 & 58.18 & 64.39 & 55.43 & 55.02 & 53.73 \\
 & ProposalContrast\cite{DBLP:conf/eccv/proposal} & 59.52 & 55.91 & 58.69 & 58.22 & 64.53 & 55.45 & 55.36 & 54.07 \\
 & {MV-JAR~\cite{mvjar}} & {62.28} & {59.15} & 61.88 & 61.45 & 66.98 & 59.02 & 57.98 & 57.00 \\ 
 &\cellcolor{Gray} BEV-MAE (Ours) &\cellcolor{Gray} \textbf{62.88}&\cellcolor{Gray}\textbf{59.97} &\cellcolor{Gray} 61.79&\cellcolor{Gray}61.37 &\cellcolor{Gray} 67.35&\cellcolor{Gray}59.39 &\cellcolor{Gray} 59.51&\cellcolor{Gray}59.14 \\
 \Xhline{0.6px}
\multirow{5}{*}{50\%} & Random & 66.43 & 63.36 & 63.81 & 63.38 & 70.78 & 63.05 & 64.71 & 63.66 \\
 & PointContrast\cite{xie2020pointcontrast} & 65.51 & 62.21 & 62.66 & 62.23 & 69.82 & 61.53 & 64.04 & 62.86 \\
 & ProposalContrast\cite{DBLP:conf/eccv/proposal} & 65.76 & 62.49 & 62.93 & 62.50 & 70.09 & 61.86 & 64.26 & 63.11 \\
 & {MV-JAR~\cite{mvjar}} & {66.70} & {63.69} & 64.30 & 63.89 & 71.14 & 63.57 & 64.65 & 63.63 \\ 
 &\cellcolor{Gray} BEV-MAE (Ours) &\cellcolor{Gray} \textbf{67.16}&\cellcolor{Gray} \textbf{64.07} &\cellcolor{Gray} 64.33&\cellcolor{Gray}63.84 &\cellcolor{Gray} 71.38&\cellcolor{Gray}63.61 &\cellcolor{Gray} 65.76&\cellcolor{Gray}64.77 \\
 \Xhline{0.6px}
\multirow{5}{*}{100\%} & Random & 68.50 & 65.54 & 64.96 & 64.56 & 72.38 & 64.89 & 68.17 & 67.17 \\
 & PointContrast\cite{xie2020pointcontrast} & 68.06 & 64.84 & 64.24 & 63.82 & 71.92 & 63.81 & 68.03 & 66.89 \\
 & ProposalContrast\cite{DBLP:conf/eccv/proposal} & 68.17 & 65.01 & 64.42 & 64.00 & 71.94 & 63.94 & 68.16 & 67.10 \\
 & {MV-JAR~\cite{mvjar}} & {69.16} & {66.20} & 65.52 & 65.12 & 72.77 & 65.28 & 69.19 & 68.20 \\ 
 &\cellcolor{Gray} BEV-MAE (Ours) &\cellcolor{Gray} \textbf{69.35} &\cellcolor{Gray} \textbf{66.46} &\cellcolor{Gray} 65.54 &\cellcolor{Gray} 65.02 &\cellcolor{Gray} 72.84 &\cellcolor{Gray} 65.31 &\cellcolor{Gray} 69.67 &\cellcolor{Gray} 69.05 \\
 \Xhline{0.8px}
\end{tabular}
\caption{
{Results about data efficiency on Waymo.} 
The detectors are fine-tuned on various fractions of Waymo training split following MV-JAR~\cite{mvjar}. `Random' denotes the training-from-scratch baseline. 
}
\label{tab:data efficiency}
\end{table*}

\section{Experimental Results}
\subsection{Implementation Details}
We evaluate the proposed BEV-MAE on two popular large-scale autonomous driving datasets, \textit{i.e.}, nuScenes and Waymo Open Dataset. 
We mainly focus on the evaluation metrics of mAP and NDS for nuScenes and the more difficult LEVEL\_2 metric (L2 mAP and L2 APH) for Waymo.
%
\footnotetext[1]{https://github.com/Nightmare-n/GD-MAE}

During pre-training, We train BEV-MAE with the Adam optimizer under the one-cycle schedule. 
The maximum learning rate is 0.0003 with a batch size of 4. 
The masking ratio is 0.7. 
The number of predicted points $L$ in point cloud reconstruction is 20 for each masked grid. 
For nuScenes, we use a common strategy~\cite{yin2020center} that concatenates one labeled point cloud frame and nine consecutive unlabeled point cloud sweeps to form a denser point cloud input.

\subsection{Waymo Results}
On the Waymo dataset, we adopt VoxelNet as the 3D encoder and use CenterPoint as the LiDAR-based object detector following ProposalContrast. 
We replace the multi-scale encoder of GD-MAE with VoxelNet following MV-JAR for fair comparisons.
\tabref{tab:main result waymo} shows that BEV-MAE substantially outperforms the training-from-scratch baselines and state-of-the-art self-supervised learning methods.
Specifically, BEV-MAE outperforms the training-from-scratch baseline by 1.42 mAP and 1.34 APH, outperforming ProposalContrast by 0.60 mAP and 0.70 APH.
BEV-MAE outperforms GA-MAE with 63\% pre-training cost.
In addition, we observe that, with only 20\% training data and 7\% computation cost, BEV-MAE achieves better results compared to ProposalContrast with 100\% data for pre-training.
Meanwhile, BEV-MAE achieves comparable results to MSP with 20\% pre-training data.
The results demonstrate the efficiency of BEV-MAE pre-training.

\subsection{NuScenes Results}
For the nuScenes dataset, we apply BEV-MAE to a strong 3D object detector, \textit{i.e.,} TransFusion-L.
\tabref{tab:main result nus} compares BEV-MAE with several LiDAR-based 3D object detectors on the nuScenes \textit{test} split. 
BEV-MAE outperforms TransFusion-L baseline by 1.5 NDS and 1.5 mAP, surpassing Link by 0.7 NDS.
Notably, GeoMAE uses a multi-stride structure to improve detection performance further. 
To compare with GeoMAE, we increase the number of channels and layers for the convolution block in the 3D encoder. 
BEV-MAE outperforms GeoMAE and achieves new state-of-the-art results with 73.6 NDS and 69.6 mAP.

\subsection{Data Efficiency}
To assess the data efficiency of BEV-MAE, we train the 3D detectors with different amounts of labeled data. 
We follow the settings of the data-efficient benchmark proposed by MV-JAR~\cite{mvjar} to split the fine-tuning dataset and train the detectors.
As shown in \tabref{tab:data efficiency}, pre-training with BEV-MAE can consistently improve the detection results on different fractions of the training data.
Especially, our method can obtain more significant gains when the labeled data is less. 
For example, BEV-MAE surpasses the training-from-scratch baseline by 7.22 mAP and 7.43 APH on L2 when using 5\% labeled data.
These results suggest the potential of BEV-MAE in using large amounts of unlabeled data.
Furthermore, we observe that BEV-MAE brings marginal improvements when fine-tuned on the full Waymo dataset. 
The reason is that the detector’s capacity becomes the bottleneck for detection performance~\cite{mvjar}.

\subsection{Transfer Learning}
We evaluate the cross-dataset transferability of the pre-trained encoder by fine-tuning it on a different dataset. 
Since the collection of different datasets uses various types of sensors, we only use the coordinates as the input feature of point clouds for compatibility across datasets.
\tabref{tab:transfer} shows that BEV-MAE consistently improves the detection performance across different datasets. 
In addition, we find that CenterPoint achieves better results when pre-training and fine-tuning on the same dataset. 
The domain gaps between two datasets likely affect the transfer performance negatively, \textit{e.g.}, the density of point clouds on Waymo is five times higher than it of nuScenes.

Moreover, we conduct a more realistic pre-training setting, \textit{i.e.}, pre-trained on a combined dataset and fine-tuned on the target dataset. 
%
The performance of the 3D object detector on Waymo and nuScenes can be further improved under this setting.
These results indicate that BEV-MAE can exploit existing various datasets for better pre-training.

\setlength\tabcolsep{.1em}
\begin{table}[!t]
\begin{center}
\small
\begin{tabular}{c|cc|cc}
\Xhline{0.8px}
\multirow{2}{*}{\diagbox{Pre-train}{Fine-tune}} & \multicolumn{2}{c|}{nuScenes} & \multicolumn{2}{c}{Waymo} \\
\cline{2-5}
& mAP & NDS & L2 mAP & L2 APH \\
\Xhline{0.6px}
Random init. & 48.6 & 58.4 & 63.97 & 61.53 \\
nuScenes & {49.7${{^{+1.1}}}$} & {58.9${{^{+0.5}}}$} & 64.79${^{+0.82}}$ & 62.28${^{+0.75}}$ \\
Waymo & {49.4${{^{+0.8}}}$} & {58.8${{^{+0.4}}}$} & 65.13${^{+1.16}}$ & 62.63${^{+1.10}}$ \\
\rowcolor{Gray}
nuScenes + Waymo & \textbf{50.1}${^{\bm{+1.5}}}$ & \textbf{59.1}${^{\bm{+0.7}}}$ & \textbf{65.36}${^{\bm{+1.39}}}$ & \textbf{62.89}${^{\bm{+1.36}}}$ \\

\Xhline{0.8px}
\end{tabular}
\caption{{Results of transfer learning.} 
The contents in the column and row show the datasets for pre-training and fine-tuning, respectively.}
\label{tab:transfer}
\end{center}
\end{table}

\setlength\tabcolsep{.2em}
\begin{table}[!t]
\begin{center}
\small
\begin{tabular}{c|cc|cc}
\Xhline{0.8px}
Pre-train & Reconstruction target & LT~ & L2 mAP & L2 APH \\
\Xhline{0.6px}
None & - & - & 65.60 & 63.21 \\
\Xhline{0.6px}
\multirow{5}{*}[-0.7ex]{BEV-MAE} & 
  Coord. (w/o norm) & $\checkmark$ & 65.66&63.09 \\
& Coord. (w norm) & $\checkmark$ & 66.20&63.71 \\
& Density & $\checkmark$ & 65.80&63.27 \\
& Number of points & $\checkmark$ & 65.32&62.88 \\
\cline{2-5}
& Coord. (w norm) + Density & $\times$ & 66.49 & 63.99\\
& Coord. (w norm) + Density & $\checkmark$ & \textbf{66.70} & \textbf{64.25}\\
\Xhline{0.8px}
\end{tabular}
\caption{{Ablation on main components.} `LT' denotes the shared learnable point token. Each component brings performance improvement for BEV-MAE.
}
\label{tab:main component}
\end{center}
\end{table}

\subsection{Ablation Study}
\label{ablation}
We conduct ablation experiments to analyze the effectiveness of each setting of BEV-MAE, including the main components and hyper-parameters.
In the following experiments, we use CenterPoint as the 3D detector. 
We first pre-train the 3D encoder on 20\% training data of Waymo with BEV-MAE and then evaluate its performance by fine-tuning with CenterPoint on the same 20\% training data. 

\subsubsection{Main Component.}
We evaluate the effectiveness of each component of the proposed BEV-MAE, as shown in \tabref{tab:main component}.
Pre-training with normalized coordinates or density prediction can improve 3D detection performance.
The detection results can be further improved by taking both coordinate and density prediction as the pre-training task, achieving 66.49 mAP and 63.99 APH.
For coordinate prediction, the normalization operation to process the coordinates of points is essential.

Replacing density prediction with the number of points prediction in Voxel-MAE~\cite{voxelmae} decreases the detection performance.
The reason is that since the number of points in a BEV grid varies from one to hundreds, the pre-training process is unstable when using the number of points prediction as the target.
Moreover, pre-training with the shared learnable token brings additional performance gain of 0.21 mAP and 0.26 APH.
%

\subsubsection{Decoder Design.}
In addition to one-layer 3$\times$3 convolution, we also test two decoders with more complex design, \textit{i.e.}, Transformer block~\cite{DBLP:conf/cvpr/mae} and Residual Conv block~\cite{he2016deep}, as shown in \tabref{tab:decoder}.
The results show that the detection performance drops with more complex decoders.
The performance decreases more when using the Transformer block as the decoder, which we assign to the different architecture between the encoder and decoder (Sparse Convolution \textit{vs.} Transformer).
In addition, we observe that the complex decoders bring additional training cost, \textit{e.g.}, Residual Conv block brings additional 0.2$\times$ training cost compared with one-layer 3$\times$3 Conv.
These results show that exploring decoder design may not be essential for BEV-MAE. A simple one-layer 3$\times$3 convolution is adequate for practice.
%

\setlength\tabcolsep{.2em}
\renewcommand\arraystretch{1}
\begin{table}[!t]
\begin{center}
\small
\begin{tabular}{lccc}
\toprule
Decoder & L2 mAP & L2 APH & Training cost \\
\midrule
One-layer 3$\times$3 Conv & \textbf{66.70} & \textbf{64.25} & \textbf{1$\times$}\\
Residual Conv block & 66.61 & 64.09 & 1.2$\times$\\
Transformer block & 65.80 & 63.26 &1.4$\times$\\
\bottomrule
\end{tabular}
\caption{{Ablation on different deocders.} One-layer 3$\times$3 Conv achieves the best results with the least training cost.}
\label{tab:decoder}
\end{center}
\end{table}

\setlength\tabcolsep{.15em}
\begin{table}[!t]
\begin{center}
\small
\begin{tabular}{lcccc}
\toprule
Masking strategy & L2 mAP & L2 APH & Memory &Training cost \\
\midrule
BEV-guided masking & \textbf{66.70} &\textbf{64.25} & \textbf{4.1G} & \textbf{1$\bm\times$}\\
Voxel masking & 66.63 & 64.16 & 12.6G & 1.4$\times$\\
\bottomrule
\end{tabular}
\caption{{Ablation on the masking strategy.} Pre-training with the BEV-guided masking strategy performs better with less GPU memory consumption and pre-training cost.}
\label{tab:mask strategy}
\end{center}
\end{table}

\subsubsection{Masking Strategy.}
We study how different masking strategies, including a simple voxel masking strategy and the proposed BEV-guided masking strategy, affect the effectiveness of representation learning.
For the voxel masking strategy, we randomly mask non-empty voxels and use sparse deconvolution layers~\cite{shi2020unet} as the decoder to recover the points in masked voxels.
In \tabref{tab:mask strategy}, we observe that the proposed BEV-guided masking strategy achieves better transfer performance on downstream 3D object detection tasks.
In addition, compared with the voxel masking strategy, the BEV-guided masking strategy significantly reduces the GPU memory consumption (4.1G \textit{vs.} 12.6G) and training costs (1$\times$ \textit{vs.} 1.4$\times$) during pre-training.

\setlength\tabcolsep{1em}
\begin{table}[!t]
\begin{center}
\small
\begin{tabular}{ccc}
\toprule
Masking ratio & L2 mAP & L2 APH \\
\midrule
50\% & 66.45 & 64.00 \\
60\% & 66.62 & 64.13 \\
70\% & \textbf{66.70} & \textbf{64.25}\\
80\% & 66.52 & 64.06 \\
\bottomrule
\end{tabular}
\caption{{Ablation on masking ratio.}
 The fine-tuning results are less sensitive to the masking ratio.
}
\label{tab:mask ratio}
\end{center}
\end{table}

\subsubsection{Masking Ratio.}
\tabref{tab:mask ratio} ablates the influence of the masking ratio.
The proposed BEV-MAE works well within a wide range of masking ratios (50\%-80\%).
The best results are achieved when the masking ratio is 70\%.



\section{Conclusions}
In this work, we address the problem of self-supervised pre-training on point clouds in autonomous driving scenarios.
We present BEV-MAE to pre-train the 3D encoder of LiDAR-based 3D object detectors.
Instead of simply masking points or voxels, we propose a BEV-guided masking strategy for better BEV representation learning and to avoid complex decoder design.
Furthermore, we introduce a shared learnable point token to maintain the receptive field size of the encoder during pre-training and fine-tuning.
By leveraging the properties of outdoor point clouds in autonomous driving scenarios, we propose point density prediction to guide the encoder to learn location information.
Experimental results show that BEV-MAE surpasses previous self-supervised learning methods in performance and pre-training efficiency.
Moreover, BEV-MAE can further boost the performance of the state-of-the-art 3D object detectors and achieve new state-of-the-art results on nuScenes.

\section{Acknowledgements}
This work was supported by National Key R\&D Program of China (Grant No. 2022ZD0160305). 

\bibliography{aaai24}

\end{document}